\documentclass{llncs}
\usepackage{capt-of}
\usepackage{booktabs,caption,fixltx2e}
\usepackage[flushleft]{threeparttable}
\usepackage{graphicx}
\usepackage{makeidx} 
\usepackage{amsmath}
\usepackage{multirow}

\usepackage{amsfonts}%
\usepackage{amssymb}%
\usepackage[english,oldcommands,plain,lined]{algorithm2e}
\usepackage{graphics}
\usepackage{subfig}
\usepackage[labelfont=bf]{caption}
\usepackage[font={small}]{caption}
\usepackage{transparent}
\usepackage{color}
\usepackage{breqn}
\usepackage{epstopdf}
\usepackage{wrapfig}

\usepackage{color, colortbl}
\definecolor{LightCyan}{rgb}{0.88,1,1}
 \usepackage{cite}
\usepackage[font=small,labelfont=bf]{caption}
\usepackage{enumitem}
\usepackage{multirow}

\begin{document}

\title{Error Corrective Boosting for Learning Fully Convolutional Networks with Limited Data} 

%
\author{Abhijit Guha Roy\inst{1,2,3}  \and  Sailesh Conjeti\inst{2} \and Debdoot Sheet\inst{3} \and Amin Katouzian\inst{4} \and Nassir Navab\inst {2,5} \and Christian Wachinger\inst{1}}
%
%
\institute{
$^1$Artificial Intelligence in Medical Imaging (AI-Med), KJP, LMU M\"{u}nchen, Germany.\\
$^2$Computer Aided Medical Procedures, Technische Universit\"{a}t M\"{u}nchen, Germany.\\
$^3$Indian Institute of Technology, Kharagpur, WB India.\\
$^4$IBM Almaden Research Center, Almaden, USA.\\
$^5$Computer Aided Medical Procedures, Johns Hopkins University, USA.}

\maketitle 
\vspace{-2mm}
\begin{abstract}

Training deep fully convolutional neural networks (F-CNNs) for semantic image segmentation requires access to abundant labeled data. 
While large datasets of unlabeled image data are available in medical applications, access to manually labeled data is very limited. 
We propose to automatically create auxiliary labels on initially unlabeled data with existing tools and to use them for pre-training. For the subsequent fine-tuning of the network with manually labeled data, we introduce error corrective boosting (ECB), which emphasizes parameter updates on classes with lower accuracy. Furthermore, we introduce SkipDeconv-Net (SD-Net), a new F-CNN architecture for brain segmentation that combines skip connections with the unpooling strategy for upsampling. 
The SD-Net addresses challenges of severe class imbalance and errors along boundaries. With application to whole-brain MRI T1 scan segmentation, we generate auxiliary labels on a large dataset with FreeSurfer and fine-tune on two datasets with manual annotations. Our results show that the inclusion of auxiliary labels and ECB  yields significant improvements. SD-Net segments a 3D scan in 7 secs in comparison to  30 hours for the closest multi-atlas segmentation method, while reaching similar performance. It also outperforms the latest state-of-the-art F-CNN models.
\end{abstract}

\section{Introduction}
\label{sec:intro}

Fully convolutional neural networks (F-CNNs) have gained high popularity for image segmentation in computer vision~\cite{longfcn2015,deconvnet2015,segnet} and biomedical imaging~\cite{Unet, Vnet}.
They directly produce a segmentation for all image pixels in an end-to-end fashion without the need of splitting the image into patches. 
F-CNNs can therefore fully exploit the image context avoiding artificial partitioning of an image, which also results in an enormous speed-up. 
Yet, training F-CNNs is challenging because each image serves as a single training sample and consequently much larger datasets with manual labels are required in comparison to patch-based approaches, where each image provides multiple patches. 
While the amount of unlabeled data rapidly grows, the access to labeled data is still limited due to the labour intense process of manual annotations. 
At the same time, the success of deep learning is mainly driven by supervised learning, while unsupervised approaches are still an active field of research. 
Data augmentation~\cite{Unet} artificially increases the training dataset by simulating different variations of the same data, but it cannot encompass all possible morphological variations.
We propose to process unlabeled data with existing automated software tools to create auxiliary labels. 
These auxiliary labels may not be comparable to manual expert annotations, however, they allow us to efficiently leverage the vast amount of initially unlabeled data for supervised pre-training of the network. We also propose to fine-tune such a pre-trained network using error corrective boosting (ECB), that selectively focuses on classes with erroneous segmentations.

\begin{figure}[t]
\centering
\begin{minipage}{\textwidth}
  \begin{minipage}[t]{0.65\textwidth}
\centering
\includegraphics[width=\textwidth]{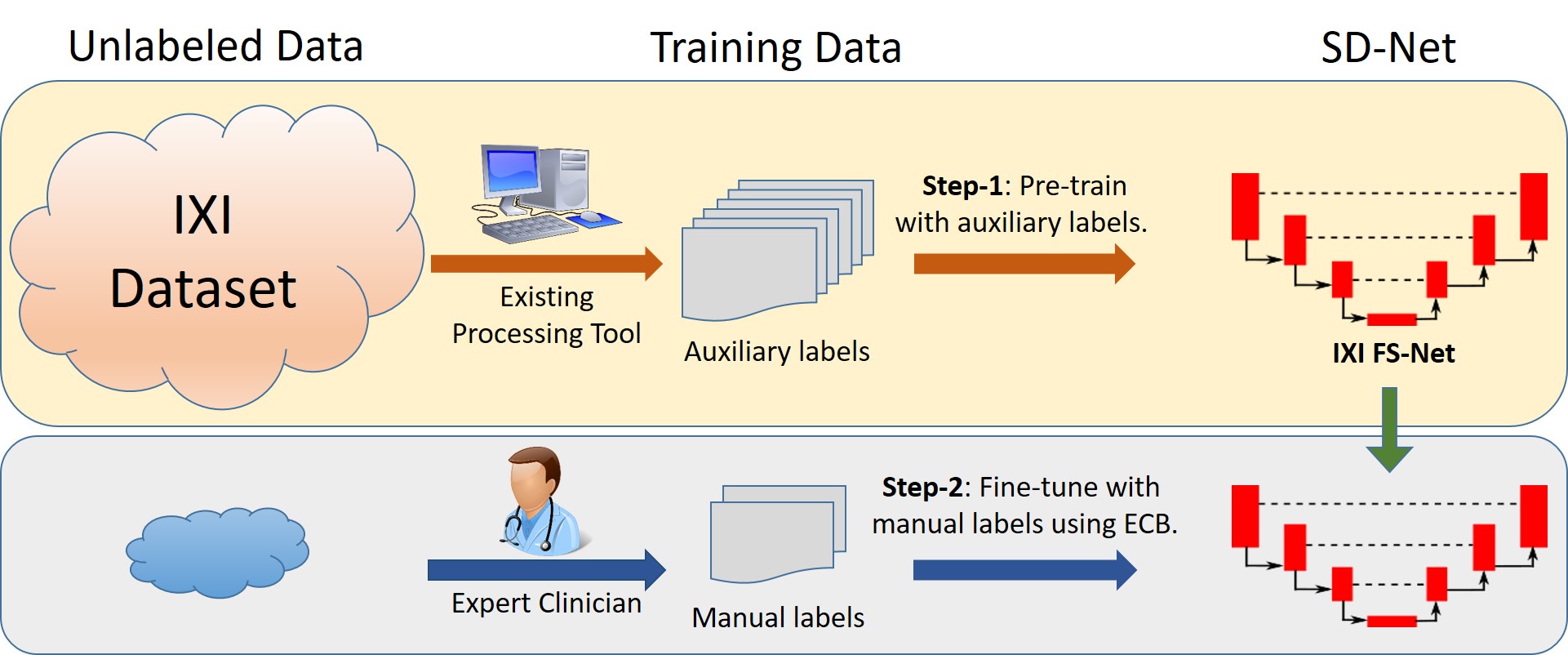}
  \end{minipage}
  \hfill
  \begin{minipage}[t]{0.33\textwidth}
\vspace{-2.4cm}
\captionof{figure}{Illustration of the different steps involved in training of F-CNNs with surplus auxiliary labeled data and limited manually labeled data.}
\label{fig:preview}
    \end{minipage}
  \end{minipage}
  \vspace{-0.7cm}
\end{figure}

In this work, we focus on whole-brain segmentation of MRI T1 scans. 
To this end, we introduce a new F-CNN architecture for segmentation, termed SkipDeconv-Net (SD-Net). 
It combines skip connections from the U-net~\cite{Unet} with the passing of indices for unpooling similar to DeconvNet~\cite{deconvnet2015}. This architecture provides rich context information while facilitating the segmentation of small structures. 
To counter the severe class imbalance problem in whole-brain segmentation, we use median frequency balancing~\cite{segnet} together with placing emphasis on the correct segmentation along anatomical boundaries.
For the creation of auxiliary labels, we segment brain scans with FreeSurfer~\cite{freesurfer}, a standard tool for automated labeling in neuroimaging. 
Fig.~\ref{fig:preview} shows the steps involved in the training process. 
First, we train SD-Net on a large amount of data with corresponding auxiliary labels, in effect creating a network that imitates FreeSurfer, referred as FS-Net.
Second, we fine-tune FS-Net with limited manually labeled data with ECB, to improve the segmentation incorrectly represented by FS-Net. 

\noindent
\textbf{Related work:}
F-CNN models have recently attracted much attention in segmentation. The FCN model~\cite{longfcn2015}  up-samples the intermediate pooled feature maps with bilinear interpolation, while the DeconvNet~\cite{deconvnet2015} up-samples with indices from the pooling layers, to reach final segmentation. 
For medical images, U-net was proposed consisting of an encoder-decoder network with skip connections~\cite{Unet}. 
For MRI T1, eight sub-cortical structures were segmented using an F-CNN model, with slices in~\cite{tsogkas} and with patches in~\cite{dolz16}.
Whole-brain segmentation with CNN using 3D patches was presented in~\cite{brebisson2015deep} and~\cite{deepNAT2017}.
To the best of our knowledge, this work is the first  F-CNN model for whole-brain segmentation. 
To address the challenge of training a deep network with limited annotations, previous works fine-tune models pre-trained for classification on natural images~\cite{transfer,finetuning}. 
In fine-tuning, the training data is replaced by data from the target application with additional task specific layers and except for varying the learning rate, the same training procedure is used. 
With ECB, we change the class specific penalty in the loss function to focus on regions with high inaccuracies. 
Furthermore, instead of relying on pre-training on natural images that exhibit substantially different image statistics and are composed of three color channels, we propose using auxiliary labels to directly pre-train an F-CNN, tailored for segmenting T1 scans.

\section{Method}
\vspace{-2.5mm}
\subsection{SD-Net for Image Segmentation}
We describe the architecture, loss function, and model learning of the proposed SD-Net for image segmentation in the following section: 

\begin{figure}[t]
\centering
\includegraphics[width=\textwidth]{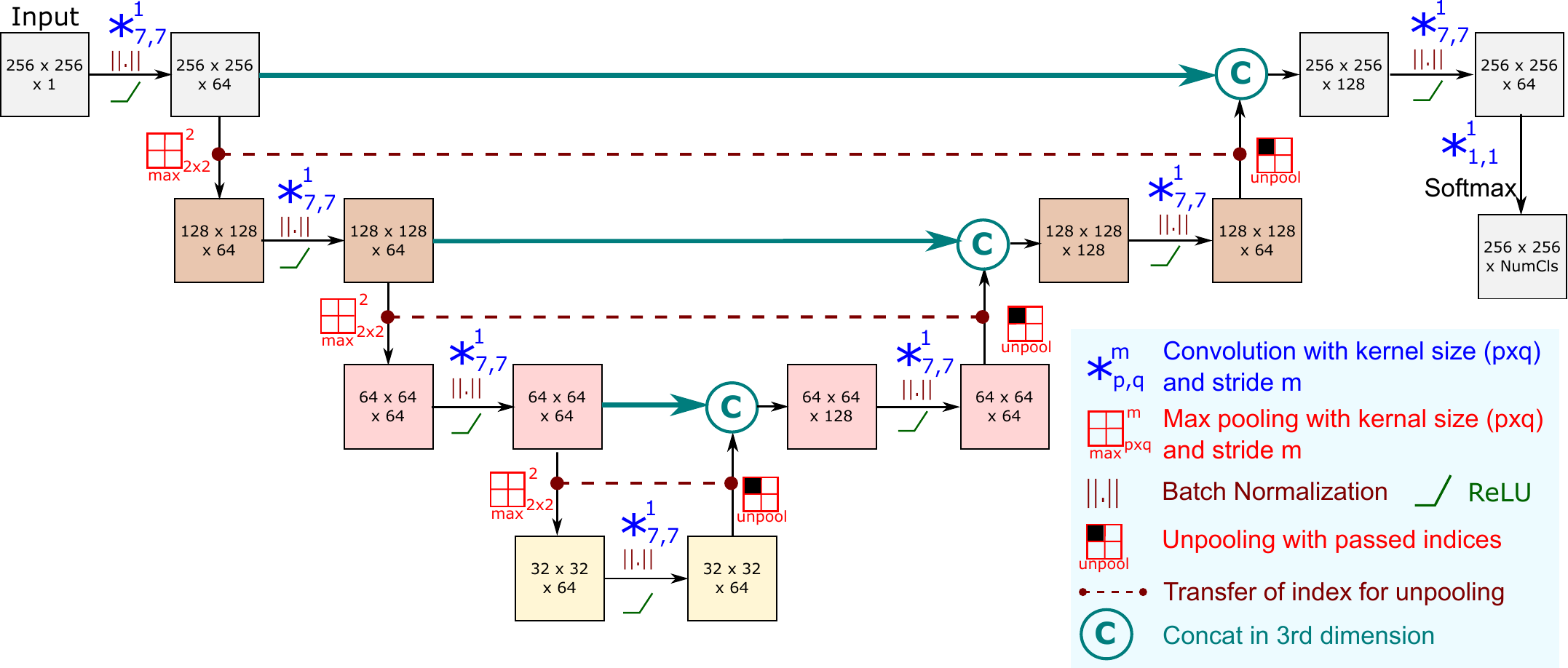}
\caption{Illustration of the proposed SkipDeconv-Net (SD-Net) architecture.}\vspace{-3mm}
\label{fig:sdnet}
\end{figure}

\noindent
\textbf{Architecture:} 
The SD-Net has an encoder-decoder based F-CNN architecture consisting of three encoder and three decoder blocks followed by a classifier with softmax to estimate probability maps for each of the classes. 
It combines skip connections from U-net~\cite{Unet} and the passing of indices for unpooling from DeconvNet~\cite{deconvnet2015}, hence the name SkipDeconv-Net (SD-Net). 
We use skip connections between the encoder and decoder as they provide rich contextual information for segmentation and also a path for the gradients to flow from the shallower decoder to the deeper encoder part during training.
In contrast to U-net where upsampling is done by convolution, we use unpooling, which offers advantages for segmenting small structures by placing the activation at the proper spatial location. 
Fig.~\ref{fig:sdnet} illustrates the  network architecture for segmenting a 2D image. 

Each encoder block consists of a $7\times7$ convolutional layer, followed by a batch normalization layer and a ReLU (Rectifier Linear Unit) activation function. Appropriate padding is provided before every convolution to ensure similar spatial dimensions of input and output. 
With the $7\times7$ kernels, we have an effective receptive field at the lowest encoder block that almost captures the entire brain mask. 
It therefore presents a good trade-off between model complexity and the capability of learning long-range connections. 
Each encoder block is followed by a max pooling layer, reducing the spatial dimension of feature maps by half.
Each decoder block consists of an unpooling layer, a concatenation by skip connection, a $7\times7$ convolutional layer, batch normalization and ReLU function. The unpooling layer upsamples the spatial dimension of the input feature map by using the saved indices with maximum activation during max pooling of the corresponding encoder block. 
The remaining locations are filled with zeros.
Unpooling does not require to estimate parameters, in contrast to the up-convolution in U-net. 
The unpooled feature maps are concatenated with the feature maps of the encoder part that have the same spatial dimension. 
The following convolution layer densifies the sparse unpooled feature maps for smooth prediction. 
The classifier consists of a $1\times1$ convolutional layer to transfer the $64$ dimensional feature map to a dimension corresponding to number of classes~($N$) followed by a softmax layer.

\noindent
\textbf{Loss Function:} SD-Net is trained by optimizing two loss functions: (i) weighted multi-class logistic loss and (ii) Dice loss. 
The logistic loss provides a probabilistic measure of similarity between the prediction and ground truth. 
The Dice loss is inspired by the Dice overlap ratio and yields a true positive count based estimate of similarity\cite{Vnet}.
Given the estimated probability $p_l(\mathbf{x})$ at pixel $\mathbf{x}$ to belong to the class $l$ and the ground truth probability $g_l(\mathbf{x})$, the loss function is 
\begin{equation}
\mathcal{L} =  \underbrace{ -\sum_{\mathbf{x}} \omega(\mathbf{x}) g_l(\mathbf{x}) \log(p_{l}(\mathbf{x}))}_{\mathrm{Logistic Loss}} - \underbrace{\frac{2 \sum_{\mathbf{x}} p_{l}(\mathbf{x}) g_{l}(\mathbf{x})}{\sum_{\mathbf{x}} p_{l}^2(\mathbf{x}) + \sum_{\mathbf{x}} g_{l}^2(\mathbf{x})}.}_{\mathrm{Dice Loss}}
\label{eq:cost}
\end{equation}
 
We introduce weights $\omega(\mathbf{x})$ to tailor the loss function to challenges that we have encountered in image segmentation: the class imbalance and the segmentation errors along anatomical boundaries. 
Given the frequency $f_l$ of class $l$ in the training data, i.e., the class probability, the indicator function $I$, the {training} segmentation~$S$, and the 2D gradient operator $\nabla$, the weights are defined as
\begin{equation}
\omega(\textbf{x}) = \sum_l I( S(\mathbf{x})==l) \ \frac{median(\mathbf{f})}{f_l} + \omega_0 \cdot I(|\nabla S(\mathbf{x})|>0)
\label{eq:weight}
\end{equation}
with the vector of all frequencies $\mathbf{f} = [f_1, \ldots, f_N]$.
The first term models median frequency balancing~\cite{segnet} and compensates for the class imbalance problem by highlighting classes with low probability. 
The second term puts higher weight on anatomical boundary regions to emphasize on the correct segmentation of contours. $\omega_0$ balances the two terms.

\noindent
\textbf{Model Learning:} We learn the SD-Net with stochastic gradient descent. The learning rate is initially set to $0.1$ and reduced by one order after every $20$ epochs till convergence. 
The weight decay is set to $0.0001$. 
Mini batches of size $10$ images are used, constrained by the $12$ GB RAM of the Tesla K40 GPU. 
A high momentum of $0.9$ is set to compensate for this small batch size. 

\vspace{-2.5mm}
\subsection{Fine-Tuning with Error Corrective Boosting}
Since the SD-Net directly predicts the segmentation of the entire 2D slice, each 3D scan only provides a limited number of slices for training. 
Due to this limited availability of manually labeled brain scans and challenges of unsupervised training, we propose to use large scale auxiliary labels for assisting in training the network. 
The auxiliary labels are created with FreeSurfer~\cite{freesurfer}. 
Although these labels cannot replace extensive manual annotations, they can be automatically computed on a large dataset and be used to train FS-Net, which is essentially an F-CNN mimicking FreeSurfer. 
To the best of our knowledge, this work is the first application of auxiliary, computer-generated labels for training neural networks for image segmentation.

Pre-training provides a strong initialization of the network and we want to use the manually labeled data to improve on brain structures that are poorly represented by the auxiliary labels. 
To this end, we introduce error corrective boosting (ECB) for fine-tuning, which boosts the learning process for  classes with high segmentation inaccuracy. 
ECB iteratively updates the weights in the logistic loss function in Eq.~(\ref{eq:cost}) during fine-tuning. 
We start the fine-tuning with the standard weights as described in Eq.~(\ref{eq:weight}). 
At epoch $t>1$, we iteratively evaluate the accuracy $a_l^t$ of class~$l$ on the validation set. 
The weights are updated for each epoch, following an approach that could be considered as median accuracy balancing as shown in Eq.~(\ref{eqn:ECB}).
\begin{equation}
\omega^{(t+1)}(\textbf{x}) = \sum_l I( S(\mathbf{x})==l) \ \frac{median(\mathbf{a}^t)-m^t}{{a_l}^t-m^t}
\label{eqn:ECB}
\end{equation}
with the vector of accuracies $\mathbf{a}^t = [a_1^t, \ldots, a_N^t]$ and the margin $m^t=\min({\mathbf{a}}^t)-q$ that normalizes the accuracies with respect to the least performing class. 
The constant $q$ is set to $0.05$, i.e. $5\%$, to avoid numerical instability. 
Error corrective boosting sets high weights for classes with low accuracy to selectively correct for errors in the auxiliary labels, which is particularly helpful for whole-brain segmentation with a large number of classes.

\vspace{-2mm}
\section{Results}
\label{sec:experiment}
\vspace{-2mm}
\noindent
\textbf{Datasets:} We pre-train the networks with FreeSurfer labels using 581 MRI-T1 volumes from the IXI dataset\footnote{http://brain-development.org/ixi-dataset/}. These volumes were acquired from 3 different hospitals with different MRI protocols. 
For the fine-tuning and validation, we use two datasets with manual labels: (i) 30 volumes from the MICCAI Multi-Atlas Labeling challenge~\cite{OasisMultiAtlas} and (ii) 20 volumes from the MindBoggle dataset~\cite{MindBoggle}. 
Both datasets are part of OASIS~\cite{oasis}. 
In the challenge dataset, $15$ volumes were used for training, $5$ for validation and $10$ for testing. In the MindBoggle dataset, $10$ volumes were used for training, $5$ for validation and $5$ for testing. 

We segment the major $26$ cortical and sub-cortical structures on the challenge data and $24$ on MindBoggle, as left/right white matter are not annotated.

\noindent
\textbf{Baselines:} 
We evaluate our two main contributions, the SD-Net architecture for segmentation and the inclusion of auxiliary labels in combination with ECB. 
We compare SD-Net to two state-of-the-art networks for semantic segmentation, U-net~\cite{Unet} and FCN~\cite{longfcn2015}, and also to a variant of SD-Net without Dice loss.
For the auxiliary labels, we report results for (i) directly deploying the IXI pre-trained network (IXI FS-Net), (ii) training only on the manually annotated data, (iii) normal fine-tuning,  and (iv) ECB-based fine-tuning. We use data augmentation with small spatial translations and rotations in all models during training.
We also compare to PICSL~\cite{picsl} (winner) and  spatial STAPLE\cite{staple} (top 5) for the challenge data whose results were available.

\begin{figure}[t]
\centering
\includegraphics[width=0.95\textwidth]{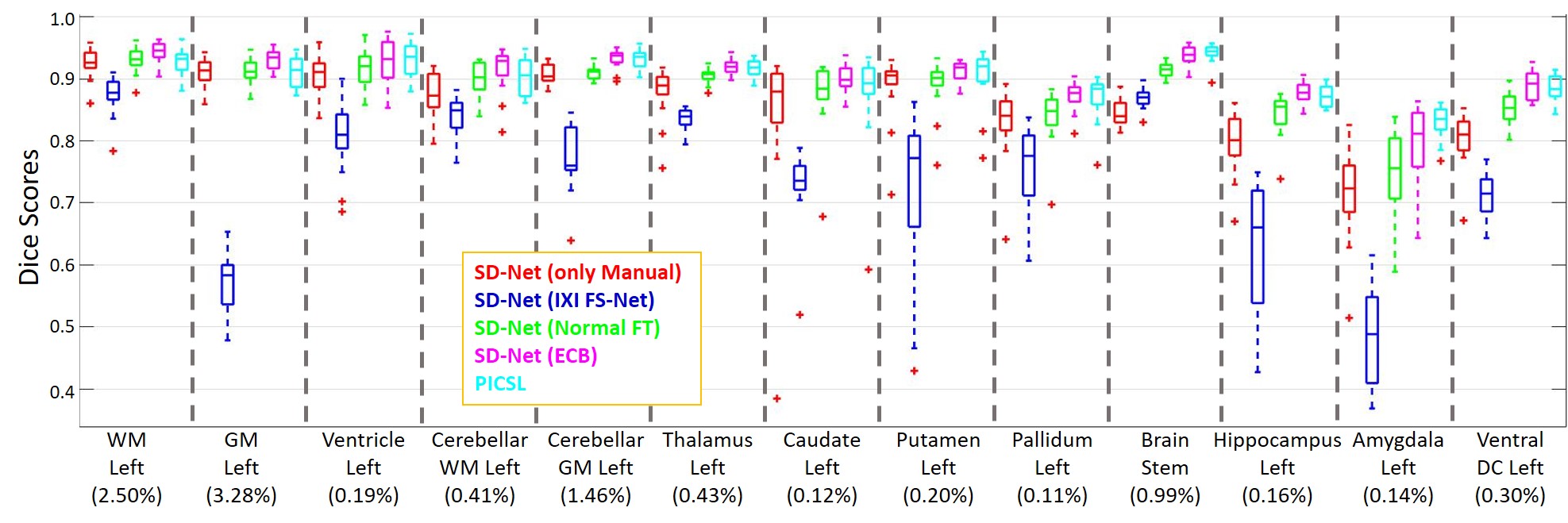} 
\vspace{-3mm}
\caption{Boxplot of Dice scores for all structures on the left hemisphere. 
Comparison of different training strategies of the SD-Net with PICSL. Class probabilities are reported to indicate the severe class imbalance, about 88\% are background.\vspace{-2mm}}
\label{fig:boxplot}
\end{figure}

\begin{table}[b]
\tiny
\vspace{-4mm}
\caption{Mean and standard deviation of the Dice scores for the different F-CNN models and training procedures on both datasets.}
  \begin{tabular}{|p{0.75in}|p{0.46in}|p{0.46in}|p{0.46in}|p{0.46in}|p{0.46in}|p{0.46in}|p{0.46in}|p{0.46in}|}
    \hline
     & \multicolumn{4}{c|}{\textbf{Multi-Atlas Challenge Dataset}} &
      \multicolumn{4}{c|}{\textbf{MindBoggle Dataset}} \\ 
     Method & IXI FS-Net & Manual Labels & Normal FT & ECB FT & IXI FS-Net & Manual Labels & Normal FT & ECB FT\\
    \hline
    \textbf{SD-Net} & $0.74\pm0.13$ & $0.85\pm0.08$ & $0.88\pm0.06$ & $\textbf{0.91}\pm0.05$ & $0.71\pm0.17$ & $0.82\pm0.06$ & $0.86\pm0.07$ & $\textbf{0.87}\pm0.06$ \\ \hline
    SD-Net (No Dice) & $0.72\pm0.14$ & $0.82\pm0.10$ & $0.84\pm0.10$ & $0.88\pm0.06$ & $0.69\pm0.10$ & $0.80\pm0.07$ & $0.85\pm0.10$ & $\textbf{0.87}\pm0.10$ \\ \hline
    U-Net\cite{Unet} & $0.71\pm0.15$ & $0.81\pm0.09$ & $0.82\pm0.11$ & $0.87\pm0.06$ & $0.69\pm0.19$ & $0.76\pm0.11$ & $0.84\pm0.07$ & $0.86\pm0.06$ \\ \hline
    FCN\cite{longfcn2015} & $0.55\pm0.23$ & $0.70\pm0.15$ & $0.78\pm0.12$ & $0.85\pm0.07$ & $0.45\pm0.24$ & $0.64\pm0.23$ & $0.81\pm0.08$ & $0.83\pm0.08$ \\ \hline
    Spatial Staple\cite{staple} & \multicolumn{4}{c|}{$0.89\pm0.05$} &
      \multicolumn{4}{c|}{NA} \\ \hline
    PICSL\cite{picsl} & \multicolumn{4}{c|}{$\textbf{0.91}\pm0.04$} &
      \multicolumn{4}{c|}{NA} \\
    \hline
  \end{tabular}
  \label{tab:res}
\end{table}

\noindent
\textbf{Results:}
Table~\ref{tab:res} lists the mean Dice scores on the test data of both datasets for all methods. 
We first compare the different F-CNN architectures, columns in the table. 
U-net outperforms FCN on all training scenarios, where the accuracy of FCN is particularly poor on the IXI FS-Net. 
The SD-Net shows the best performance with an average increase of $2\%$ mean Dice score over U-Net, significant with $p<0.01$. 
The SD-Net without the Dice loss in Eq.~(\ref{eq:cost}) does not perform as well as the combined loss. We also retrained SD-Net with only limited manual annotated data with $\omega_0 = \{3, 4, 5, 6, 7\}$, resulting in the respective mean dice scores $\{0.85, 0.83, 0.85, 0.84, 0.85\}$. These results show that there is a limited sensitivity to $\omega_0$ and we set it to $5$ for the remaining experiments.

Next, we compare the results for the different training setups, presented as rows in the table. 
Training on the FreeSurfer segmentations of the IXI data yields the worst performance, as it only includes the auxiliary labels. 
Importantly, fine-tuning the FS-Net with the manually labeled data yields a substantial improvement over directly training from the manual labels. 
This confirms the advantage of initializing the network with auxiliary labels. 
Moreover, ECB fine-tuning leads to further improvement of the Dice score in comparison to normal fine-tuning. 
On the challenge dataset, this improvement is statistically significant with $p<0.01$. 
Finally, SD-Net with ECB results in significantly higher Dice scores ($p=0.02$) than spatial STAPLE and the same Dice score as PICSL. 

\begin{figure}[t]
\centering
\includegraphics[width=0.85\textwidth]{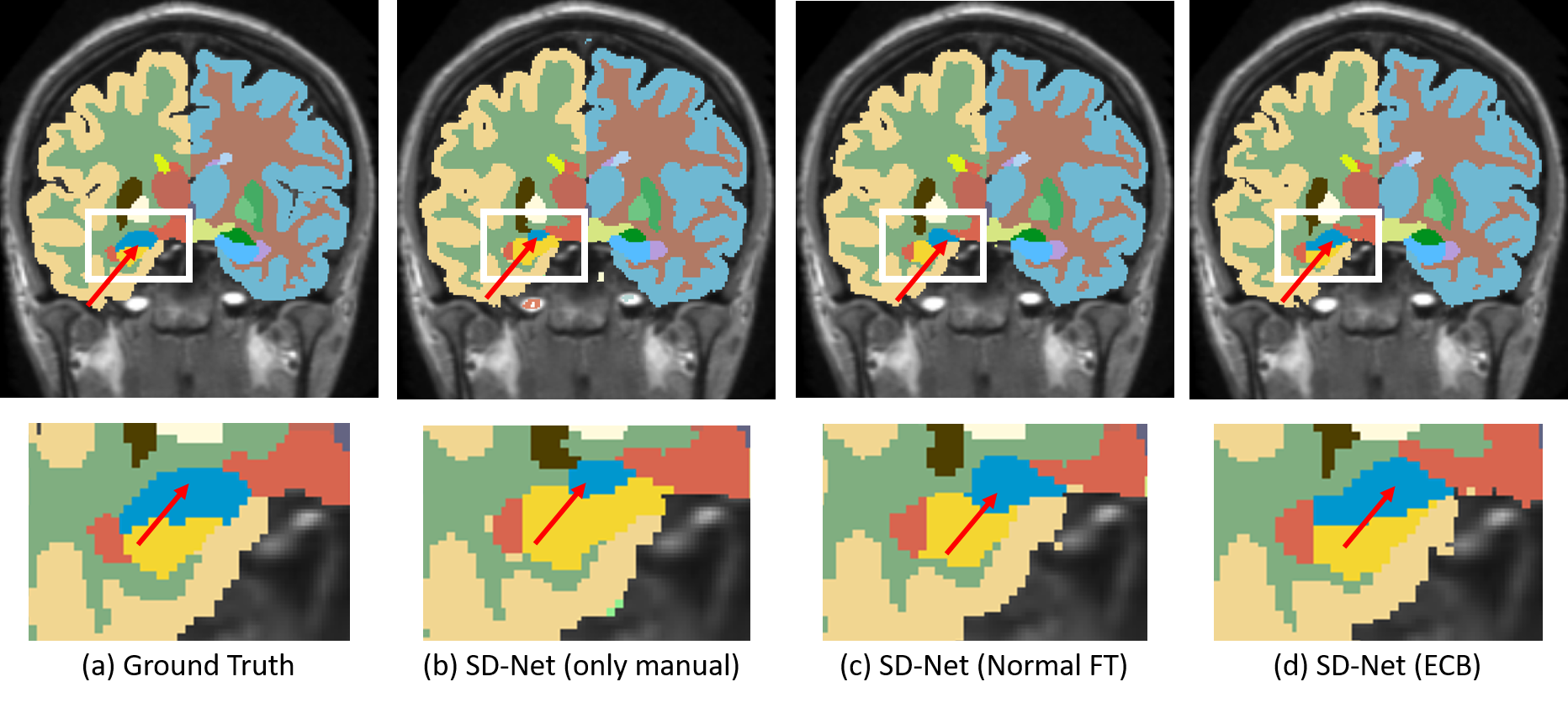}
\vspace{-4mm}
\caption{Comparison of training the SD-Net with only manual labels, normal fine-tuning and ECB together with the ground truth segmentation. 
A zoomed view of the white box is presented below, where the hippocampus (blue) is indicated by a red arrow.\vspace{-3mm}}
\label{fig:results}
\end{figure}

Fig.~\ref{fig:boxplot} presents a structure-wise comparison of the different training strategies for the SD-Net together with PICSL. The class probability for each of these structures are also presented to indicate the severe class imbalance problem.
There is a consistent increase in Dice scores for all the structures, from training with manually annotated data over normal fine-tuning to ECB. 
The increase is strongest for structures that are improperly segmented like the hippocampus and amygdala, as they are assigned the highest weights in ECB. 
Fig.~\ref{fig:results} illustrates the ground-truth segmentation together with results from the variations of training the SD-Net. 
Zoomed in regions are presented for the hippocampus, to highlight the effect of the fine-tuning. 
The hippocampus with class probability $0.16\%$ is under-segmented when trained with only limited manual data. 
The segmentation improves after normal fine-tuning, with the best results for ECB. 

Segmenting all  2D slices in a 3D volume with SD-Net takes 7 seconds on the GPU. 
This is orders of magnitude faster than multi-atlas approaches, e.g., PICSL and STAPLE, that require about 30 hours with 2 hours per pair-wise registration. 
SD-Net is also much faster than the 2-3 minutes reported for the segmentation of eight structures by the patch-based technique in~\cite{dolz16}.

\vspace{-3mm}
\section{Conclusion}
\vspace{-2.5mm}

We introduced SD-Net, an F-CNN, encoder-decoder architecture with unpooling that jointly optimizes logistic and Dice loss. 
We proposed a training strategy with limited labeled data, where we generated auxiliary segmentations from unlabeled data and fine-tuned the pre-trained network with ECB. 
We demonstrated  that (i) SD-Net outperforms U-net and FCN, (ii) using auxiliary labels improves the accuracy and (iii) ECB exploits the manually labeled data better than normal fine-tuning.
Our approach achieves state-of-the-art performance for whole-brain segmentation while being orders of magnitude faster.

\noindent
\textbf{Acknowledgement:} This work was supported in part by the Faculty of Medicine at LMU (F\"{o}FoLe), the Bavarian State Ministry of Education, Science and the Arts in the framework of the Centre Digitisation.Bavaria (ZD.B), the NVIDIA corporation and DAAD (German Academic Exchange Service). The authors would also like to thank Magdalini Paschali for proof reading and feedback.

\end{document}